# A Simple Method to improve Initialization Robustness for Active Contours driven by Local Region Fitting Energy

**Keyan Ding**, School of Mechanical and Electric Engineering, Soochow University, Suzhou 215021, People's Republic of China. Email: dingkeyan93@outlook.com

**Linfang Xiao**, College of Biosystems Engineering and Food Science, Zhejiang University, Hangzhou 310058, People's Republic of China.

**Abstract:** Active contour models based on local region fitting energy can segment images with intensity inhomogeneity effectively, but their segmentation results are easy to error if the initial contour is inappropriate. In this paper, we present a simple and universal method of improving the robustness of initial contour for these local fitting-based models. The core idea of proposed method is exchanging the fitting values on the two sides of contour, so that the fitting values inside the contour are always larger (or smaller) than the values outside the contour in the process of curve evolution. In this way, the whole curve will evolve along the inner (or outer) boundaries of object, and less likely to be stuck in the object or background. Experimental results have proved that using the proposed method can enhance the robustness of initial contour and meanwhile keep the original advantages in the local fitting-based models.

## 1. Introduction

Active contour models have been widely applied in image segmentation field since the presentation by Kass *et al.* [1] Existing active contour models can be roughly categorized into two basic classes: edge-based models [2-5] and region-based models [6-12]. Edge-based models often use an edge indicator to drive the curve towards the object boundaries, such as geodesic active contour (GAC) model [2, 3]. Region-based models usually use a certain region descriptor to find a partition on the image domain, such as the Chan–Vese (CV) model [7], but it cannot work well for images with intensity inhomogeneity because it only utilizes the image intensities from the perspective of global image.

In order to efficiently handle the intensity inhomogeneity which is often occurred in real images, Li *et al.* [8] presented an active contour model based on region-scalable fitting (RSF) energy. The RSF model draws upon the local image information by a kernel function. With the information of local image intensities, the RSF model can segment images with intensity inhomogeneity effectively. But its segmentation result largely depends on the initial contour. When the initial contour is set inappropriately, the RSF model will be stuck in local minima because the energy functional is non-convex. That means an improper initial contour will lead to a wrong segmentation result.

Zhang *et al.* [9] presented an active contour model driven by local image fitting (LIF) energy. By extracting the local image information, it is able to segment images with intensity inhomogeneity. Compared to the RSF model, the segmentation efficiency of LIF model is higher because only two convolutions are computed at each iteration while there are at least four convolutions in the RSF model. However, the problem of initialization is not solved. Liu *et al.* [10] proposed a local region-based Chan–Vese (LRCV) model. Similarly, it can segment images with intensity inhomogeneity, and the segmentation efficiency is higher than RSF model. But it is also sensitive to initialization. Wang *et al.* [11] presented an active contour model based on local Gaussian distribution fitting (LGDF) energy. It defines a local Gaussian distribution fitting energy by using local means and variances as variables. The LGDF model can distinguish regions with similar intensity means but different variances. But the problem of initialization is still unsolved and the segmentation efficiency is relatively low due to the extra computation of variances. Considering that the object and background in many real-world images are hard to be described by a predefined distribution, Liu *et al.* [12] proposed a nonparametric active contour model driven by a local histogram fitting (LHF) energy. It defines two fitting histograms that approximate the distribution of object and background locally. The LHF model can segment the regions whose distribution is hard to be predefined. However, it has low segmentation efficiency because the histogram distribution of each grey value needs to be calculated. Likewise, it is also sensitive to initialization. In addition, He *et al.* [13] proposed a local entropy-based weighted RSF model. Wang *et al.* [14] presented a local and global Gaussian distribution fitting (LGGDF) model based on LGDF model. Ji *et al.* [15] proposed a local likelihood image fitting (LLIF) model based on LGDF model and LIF model. These models are also sensitive to initial contour to some extent.

In summary, active contour models based on local region fitting energy in [8–15] can segment images with intensity inhomogeneity effectively. But an inappropriate initial contour can reduce the segmentation efficiency greatly, and even cause the failure of segmentation. It is difficult to ensure that user can find a suitable initial contour quickly. Therefore, it is important to find an effective way to address the initialization problem in the local fitting-based models. The popular methods to improve the robustness of initialization mainly include:

(1) Using the results of other segmentation methods, such as OTSU method, K-means and FCM clustering, as the initial contour of active contour model automatically. For example, Bhadauria *et al.* [16] used the results of FCM clustering to initialize the RSF model. Similarly, Gupta *et al.* [17] used the results of Gaussian kernel induced fuzzy C-means (GKFCM) clustering to initialize the RSF model. This kind of approach may cause a wrong segmentation because it is difficult to ensure that the every pre-segmentation results are desirable for different images.

(2) Combining other robust energies with local fitting energy by using a weight coefficient, i.e., the local and global intensity fitting (LGIF) model proposed by Wang *et al.* [18]. (The LGIF energy is defined by a linear combination of CV and RSF energy). In addition, Ding *et al.* [19] proposed a model driven by RSF energy and optimized LoG energy. Luo *et al.* [20] integrated image gradient alignment term into CV and RSF energy. This kind of approach can relatively improve the robustness against initialization, but the results largely depend on the value of weight coefficient, and the costs of computation are increased.

(3) Transforming non-convex function into global convex function, i.e., the global convex method proposed by Chan *et al.* [21]. In this way, the model energy would never be stuck in local minima. In addition, Yang *et al.* [22] utilized global convex segmentation method and Split Bregman method to minimize the RSF energy, thus, the proposed model is insensitive to initial contour. This kind of method is well supported by mathematical principles, but its algorithm is somewhat complicated.

In this paper, we propose a simple, universal and efficient method to improve the robustness of initialization for the active contour models based on local region fitting energy. First, we take the RSF model as an example to analyze the reason why these models are sensitive to initialization from the perspective of the curve evolution. Next, we present the proposed method and apply it to the local fitting-based models. The main idea of the proposed method is exchanging the fitting values on the two sides of contour at each iteration, which is realized by using *Minimum* and *Maximum* functions. Thus, the whole curve will evolve along the inner (or outer) boundaries of object. As a result, the curve is less likely to be stuck in the object or background. Experimental results of improved models have showed that using the proposed method can enhance the robustness of initial contour.

The remainder of this paper is organized as follows. Section 2 briefly reviews the RSF, LIF and LGDF models. Section 3 introduces the proposed method. Section 4 presents the comparsion between original and improved models. Section 5 shows some discussions about the proposed method. Section 6 concludes this paper.

## 2. Previous works

### 2.1 Region-Scalable Fitting Model

Li *et al.* [8] proposed a region-scalable fitting (RSF) model for segmenting images with intensity inhomogeneity. They defined the following energy functional:

$$E^{RSF}(\phi, f_1, f_2) = \lambda_1 \int_\Omega (\int_\Omega K_\sigma(x-y) | I(y) - f_1(x)|^2 H_\varepsilon(\phi(y)) dy) dx \\ + \lambda_2 \int_\Omega (\int_\Omega K_\sigma(x-y) | I(y) - f_2(x)|^2 [1 - H_\varepsilon(\phi(y))] dy) dx \\ + \upsilon \int_\Omega \delta_\varepsilon(\phi(x)) |\nabla \phi(x)| dx + \mu \int_\Omega \frac{1}{2} (|\nabla \phi(x)| - 1)^2 dx \quad (1)$$

where, $x, y \in \Omega$, $\lambda_1$, $\lambda_2$, $\nu$ and $\mu$ are positive constants. $K_\sigma$ is a Gaussian kernel function with standard deviation $\sigma$. $f_1(x)$ and $f_2(x)$ are two smooth functions that approximate the intensities of image outside and inside the contour $C$ in a local region, respectively. $H_\varepsilon(x)$ and $\delta_\varepsilon(x)$ are regularized Heaviside and Dirac function defined by

$$\begin{cases} H_\varepsilon(x) = \frac{1}{2}(1 + \frac{2}{\pi} \arctan(\frac{x}{\varepsilon})) \\ \delta_\varepsilon(x) = \frac{\varepsilon}{\pi(\varepsilon^2 + x^2)} \end{cases} \quad (2)$$

The method of steepest descent is used to minimize the energy functional (1). Keeping level set function $\phi$ fixed and minimizing $E^{RSF}$ with respect to $f_1$ and $f_2$, the following formulations can be obtained:

$$\begin{cases} f_1(x) = \frac{\int_\Omega K_\sigma(x-y)[H_\varepsilon(\phi(y)) \cdot I(y)] dy}{\int_\Omega K_\sigma(x-y) H_\varepsilon(\phi(y)) dy} \\ f_2(x) = \frac{\int_\Omega K_\sigma(x-y)[(1 - H_\varepsilon(\phi(y)) \cdot I(y)] dy}{\int_\Omega K_\sigma(x-y)[1 - H_\varepsilon(\phi(x))] dy} \end{cases} \quad (3)$$

Keeping $f_1$ and $f_2$ fixed and minimizing $E^{RSF}$ with respect to the level set function $\phi$, the following gradient descent flow can be obtained:

$$\frac{\partial \phi}{\partial t} = -\delta_\varepsilon(\phi)(\lambda_1 e_1 - \lambda_2 e_2) \\ + \nu \delta_\varepsilon(\phi) div(\frac{\nabla \phi}{|\nabla \phi|}) + \mu(\nabla^2 \phi - div(\frac{\nabla \phi}{|\nabla \phi|})) \quad (4)$$

where $e_1$ and $e_2$ are

$$\begin{cases} e_1(x) = \int_\Omega K_\sigma(y-x) | I(x) - f_1(y)|^2 dy \\ e_2(x) = \int_\Omega K_\sigma(y-x) | I(x) - f_2(y)|^2 dy \end{cases} \quad (5)$$

### 2.2 Local Image Fitting Model

Zhang *et al.* [9] presented an active contour model driven by local image fitting (LIF) energy. This energy functional is defined by minimizing the difference between the fitted image and the original image:

$$E^{LIF}(\phi, m_1, m_2) = \frac{1}{2} \int_\Omega |I(x) - I^{fit}(x)|^2 dx \quad (6)$$

where $I^{fit}$ is the local fitting image defined as follows:

$$I^{fit}(x) = m_1(x) H_\varepsilon(\phi(x)) + m_2(x)[1 - H_\varepsilon(\phi(x))] \quad (7)$$

with

$$\begin{cases} m_1(x) = mean(I(x) : x \in \{\phi(x) < 0\} \cap \Omega_w(x)) \\ m_2(x) = mean(I(x) : x \in \{\phi(x) > 0\} \cap \Omega_w(x)) \end{cases} \quad (8)$$

where $\Omega_w$ is a truncated Gaussian window whose size is $w \times w$ and the standard deviation is $\sigma$. The means $m_1(x)$ and $m_2(x)$ can be seen as the weighted averages of the image intensities in a Gaussian window outside and inside the contour, respectively. Thus, $m_1(x)$ and $m_2(x)$ are totally the same to $f_1(x)$ and $f_2(x)$ in the RSF model. Similarly, the steepest descent method is used to minimize the energy functional (6). Note, the details of the minimization of LIF energy, as well as the following LGDF energy, will not be included in this paper.

### 2.3 Local Gaussian distribution fitting model

Wang *et al.* [11] presented an active contour model based on local Gaussian distribution fitting (LGDF) energy. It defines a local Gaussian distribution fitting energy by using local means and variances as variables:

$$E^{LGDF}(\phi, p_{1,x}, p_{2,x})$$
$$= \lambda_1 \int_\Omega (\int_\Omega K_\sigma(x-y)\log p_{1,x}(I(y))H_\varepsilon(\phi(y))dy)dx \quad (9)$$
$$+ \lambda_2 \int_\Omega (\int_\Omega K_\sigma(x-y)\log p_{2,x}(I(y))[1-H_\varepsilon(\phi(y))]dy)dx$$

where $p_{1,x}$ and $p_{2,x}$ are

$$p_{i,x}(I(y)) = \frac{1}{\sqrt{2\pi}\sigma_i(x)}\exp(-\frac{u_i(x)-I(y)}{2\sigma_i^2(x)}), \quad i=1,2 \quad (10)$$

where $u_i(x)$ and $\sigma_i^2(x)$ are local means and standard deviations of intensity, respectively. The value of $u_1(x)$ and $u_2(x)$ can be seen as the weighted averages of the image intensities in a Gaussian window outside and inside the contour, respectively. Thus, $u_1(x)$ and $u_2(x)$ are the same to $f_1(x)$ and $f_2(x)$ in the RSF model. $\sigma_1^2(x)$ and $\sigma_2^2(x)$ can be seen as the weighted variances of the image intensities in a Gaussian window outside and inside the contour, respectively. Thus, $\sigma_1^2(x)$ and $\sigma_2^2(x)$ are

$$\begin{cases} \sigma_1^2(x) = \dfrac{\int_\Omega K_\sigma(x-y)[H_\varepsilon(\phi(y))\cdot(u_1(x)-I(y))^2]dy}{\int_\Omega K_\sigma(x-y)H_\varepsilon(\phi(y))dy} \\ \sigma_2^2(x) = \dfrac{\int_\Omega K_\sigma(x-y)[(1-H_\varepsilon(\phi(y)))\cdot(u_1(x)-I(y))^2]dy}{\int_\Omega K_\sigma(x-y)(1-H_\varepsilon(\phi(y)))dy} \end{cases} \quad (11)$$

### 3 Proposed method

This section presents a simple method of improving the robustness against initial contour for the models driven by local region fitting energy.

#### 3.1 Analysis of curve evolution

First, we take the RSF model as an example to analyze the reason why local fitting-based models are sensitive to initialization from the perspective of the curve evolution. Fig. 1 shows a typical wrong segmentation result (plotted by red line) of original RSF model caused by an inappropriate initial contour (plotted by green line). The blue arrows represent the direction of curve evolution at that time. After 15 iterations (Fig. 1$b$), a part of curve $C_1$ located inside the object is evolving along the inner boundary, but other part of curve $C_2$ located outside the object is evolving along the outer boundary. Thus, the curves $C_1$ and $C_2$ cause a mutual repulsion due to the opposite direction of curve evolution. After 50 iterations (Fig. 1$c$), another curve $C_3$ that located outside the object is emerging and evolving along the outer boundary. The curve $C_2$ and $C_3$ will merge together because of the same evolution direction. On the contrary, the curve $C_1$ and $C_3$ will repulse each other. Finally, although these curves have covered all the boundaries of object, some excess curves stuck in both the foreground and background regions, as shown in Fig. 1(d). Fig. 1(e) shows the change of RSF energy $E^{RSF}$ with the increase of iterations. It proves that $E^{RSF}$ has stuck in local minima after 500 iterations.

In short, an inappropriate position of initial contour will cause that a part of curve evolves along the inner boundary and other part of curve evolves along the outer boundary. It is a process of mutual repulsion, not a process of mutual fusion, due to the opposite evolution direction. Ideally, every part of curve should be evolved along the inner (or outer) boundary, so that the whole curve can merge each other. In the next section, we will present a method to ensure the whole curve evolving with the same direction.

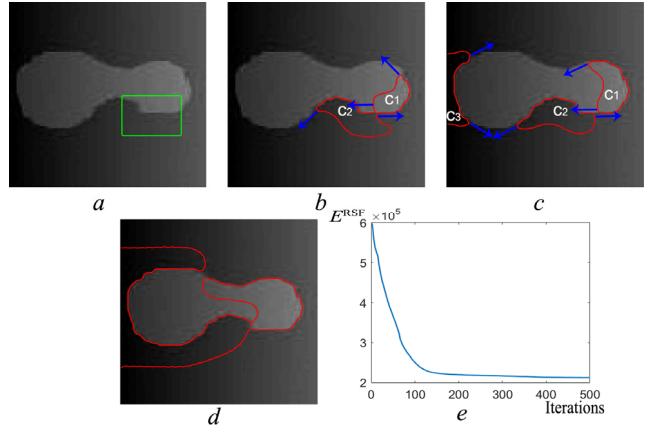

**Fig. 1.** A*nalysis of curve evolution of the RSF model.*
*a* Initial contour.
*b* After 15 iterations.
*c* After 50 iterations.
*d* After 500 iterations.
*e* The change of RSF energy with the increase of iterations.

#### 3.2 Correct the direction of curve evolution

Assuming that the whole curve evolved along the inner boundary is desirable, thus we need to reverse the curves evolving along the outer boundary. There is a key point that the values of fitting functions $f_1$ and $f_2$ are opposite when the curve evolution direction is opposite in a local region. That means when the curve is evolving along the inner boundary, the values of $f_1$ and $f_2$ are equal to the values of $f_2$ and $f_1$ when evolving along the outer boundary, respectively. Note, $f_1$ is a smooth function that fitting the image intensities outside the contour, and $f_2$ is fitting the image intensities inside the contour.

For the image with bright object and dark background, i.e., the image in Fig. 1, when the curve is evolving along the inner boundary, the value of $f_1$ should be less than $f_2$ near the boundaries according to the definition of $f_1$ and $f_2$. In the process of curve evolution, if $f_1 > f_2$ on a certain region, it means the curve is evolving along the outer boundary, i.e., the curve $C_2$ in Fig. 1$b$. Fig. 2 shows the values of $f_1$ and $f_2$ corresponding to Fig. 1$b$. In order to avoid $f_1 > f_2$ on a certain region, we can exchange the value of $f_1$ and $f_2$ in this region, i.e., the white region in Fig. 2$c$. It can be realized by using the mathematical *min* and *max* functions. Thus, we use $min(f_1, f_2)$ and $max(f_1, f_2)$ to replace $f_1$ and $f_2$, respectively.

$$\begin{cases} f_1^*(x) = \min(f_1(x), f_2(x)) \\ f_2^*(x) = \max(f_1(x), f_2(x)) \end{cases} \quad (12)$$

The gradient descent flow of RSF model can be rewritten as:

$$\frac{\partial \phi}{\partial t} = -\delta_\varepsilon(\phi)(\lambda_1 e_1^* - \lambda_2 e_2^*) \quad (13)$$
$$+ \nu \delta_\varepsilon(\phi) div(\frac{\nabla \phi}{|\nabla \phi|}) + \mu(\nabla^2 \phi - div(\frac{\nabla \phi}{|\nabla \phi|}))$$

where $e_1^*$ and $e_2^*$ are

$$\begin{cases} e_1^*(x) = \int_\Omega K_\sigma(y-x)|I(x)-f_1^*(y)|^2 dy \\ e_2^*(x) = \int_\Omega K_\sigma(y-x)|I(x)-f_2^*(y)|^2 dy \end{cases} \quad (14)$$

Note, the variables and parameters in the original RSF model are totally unchanged except the fitting functions $f_1$ and $f_2$. In this way, the whole curve will evolve along the inner boundary, and a correct segmentation result will be obtained.

On the contrary, for the image with dark object and bright background, the whole curve will evolve along the outer boundary when using above method, and a correct result will also be obtained. If we still want the whole curve evolves along the inner boundary, we can exchange the values of $f_1$ and $f_2$ in the regions where $f_1 < f_2$ to guarantee $f_1 > f_2$ at each iteration, namely using $max(f_1, f_2)$ and $min(f_1, f_2)$ to replace $f_1$ and $f_2$, respectively.

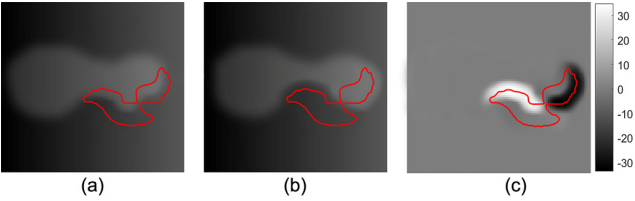

(a)      (b)      (c)

**Fig. 2.** *The value of $f_1$, $f_2$ and $f_1-f_2$ corresponding to Fig. 1b.*
*a* The value of $f_1$.
*b* The value of $f_2$.
*c* The value of $f_1-f_2$.

### 3.3 Extensions of proposed method

The above proposed method can be easily applied to other local fitting-based models. In the LIF model, because the fitting functions $m_1$ and $m_2$ are totally the same to $f_1$ and $f_2$ in the RSF model, we can use $min(m_1, m_2)$ and $max(m_1, m_2)$ (or $max(m_1, m_2)$ and $min(m_1, m_2)$) to replace $m_1$ and $m_2$, respectively, and keep the others unchanged. As a result, the improved LIF model will more robust against initial contour. Similarly, in the LGDF model, the local means of image intensities $u_1$ and $u_2$ can be replaced by $min(u_1, u_2)$ and $max(u_1, u_2)$ (or $max(u_1, u_2)$ and $min(u_1, u_2)$), respectively. At the same time, the standard deviations $\sigma_1^2$ and $\sigma_2^2$ are replaced by $min(\sigma_1^2, \sigma_2^2)$ and $max(\sigma_1^2, \sigma_2^2)$ (or $max(\sigma_1^2, \sigma_2^2)$ and $min(\sigma_1^2, \sigma_2^2)$), respectively. In the LHF model, the fitting histograms $P_1^x(z)$ and $P_2^x(z)$ can be replaced by $min(P_1^x(z), P_2^x(z))$ and $max(P_1^x(z), P_2^x(z))$ (or $max(P_1^x(z), P_2^x(z))$ and $min(P_1^x(z), P_2^x(z))$), respectively.

Besides the value of means, deviations and histograms on the two sides of curve, the proposed method can be applied to the models based on other local statistical information. In addition, there are many models used the fitting functions $f_1$ and $f_2$ in (3), such as the LRCV model [10], the LGIF model [18], the local entropy-based weighted RSF model [13], the LoG-based RSF model [19], the LGGDF model [14], the LLIF model [15], the normalized LIF model [23] and the patch-statistical region fitting model [24]. Our proposed method can be also applied to these models to enhance the robustness of initial contour.

### 4 Experimental results

In this section, we will demonstrate the effectiveness of proposed method by comparing improved models and original models, including the RSF model, the LIF model and the LGDF model. The implementation scheme of the improved models is the same as the original models totally. Moreover, the values of parameters are also the same. Each initial level set function $\phi_0$ is initialized as a binary step function which takes $-c_0$ inside zero level set and $c_0$ outside.

Unless otherwise specified, we use the following parameters: Both in the original and improved RSF model: $c_0=2$, $\sigma=3$, $\varepsilon=1$, $\lambda_1=1$, $\lambda_2=1$, $u=1$, $\nu=0.001\times255^2$ and time step $\Delta t=0.1$. Both in the original and improved LIF model: $c_0=2$, $\sigma=3$, $\varepsilon=1$, $\Delta t=0.01$, the size of regularized Gaussian kernel is $5\times5$, and its variance is 0.5. Both in the original and improved LGDF model: $c_0=2$, $\sigma=3$, $\varepsilon=1$, $\lambda_1=1$, $\lambda_2=1$, $u=0.01$, $\nu=1$ and $\Delta t=1$. The Matlab code can be downloaded at https://www.researchgate.net/publication/323748221_Robust_Initialization_of_ACM_the_matlab_code_of_the_paper.

### 4.1 Curve evolution in the improved model

First, we set an inappropriate initial contour, which will cause a wrong segmentation results in the original RSF, LIF and LGDF models. Fig. 3 shows the process of curve evolution in these original models and improved models when the initial contour is inappropriate. By comparing the curve evolution of original models and improved models for two images in Fig. 3, we notice that the proposed method can correct the evolution direction of the curve and make it evolves along the inner boundary. As a result, all of the improved models obtain a correct segmentation result.

### 4.2 Comparisons on different initial contours

Next, we set four different initial contours in the original and improved RSF, LIF and LGDF models, and make a comparison between their segmentation results. Fig. 4 shows the segmentation results of original RSF model and improved RSF model under these different initial contours. For the left image in Fig. 4, the original RSF model gets a correct result only under the first initial contour. In the improved RSF model, the desired segmentation results can be obtained with each initial contour. Note, the weight of length term $\nu=0.001\times255^2$ for this image. Similarly, for the middle image in Fig. 4, the original RSF model obtains a correct result only under the first initial contour while the improved RSF model can obtain the correct result with each initial contour. For the right image in Fig. 4, the results are also similar. Note, $\lambda_2 = 2$ in this image.

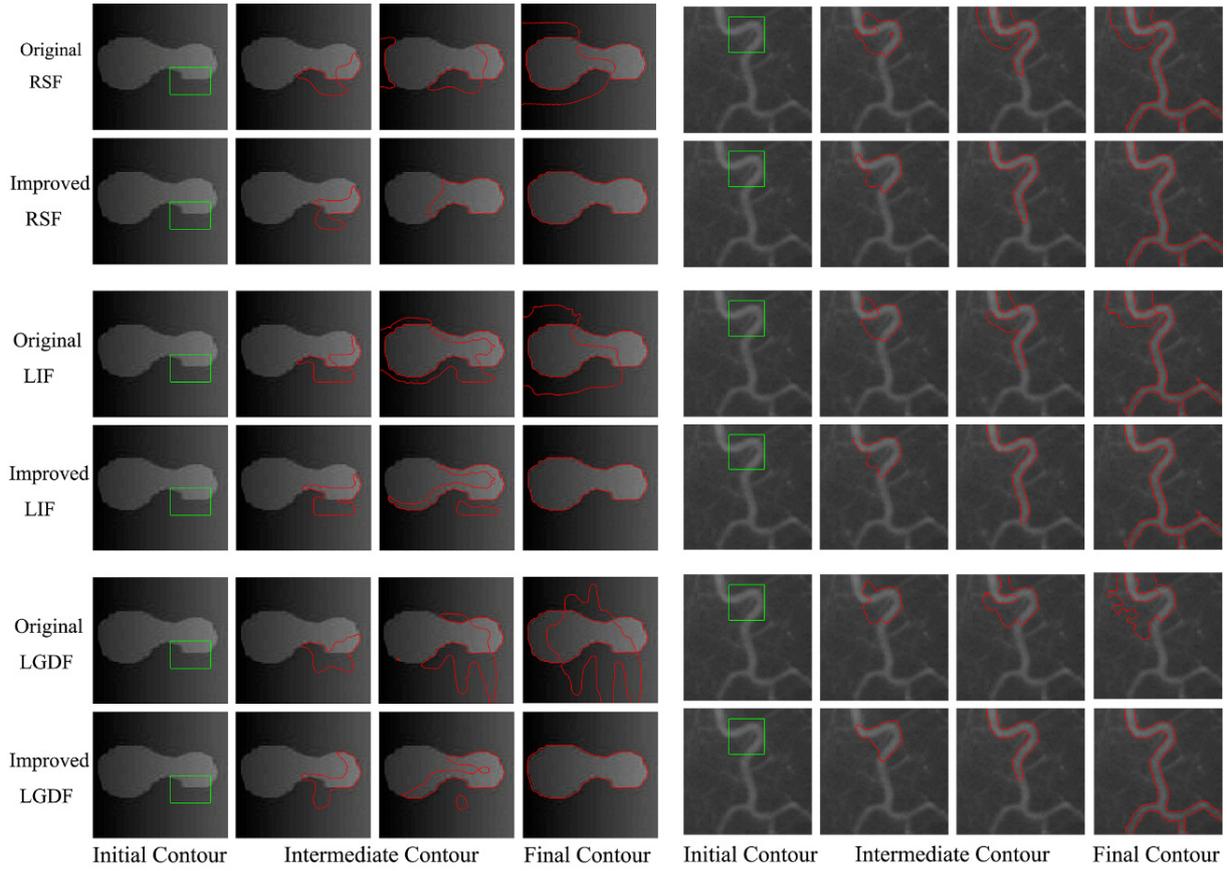

**Fig. 3** *The curve evolution process of original RSF, LIF, LGDF models and improved RSF, LIF, LGDF models for two images.*

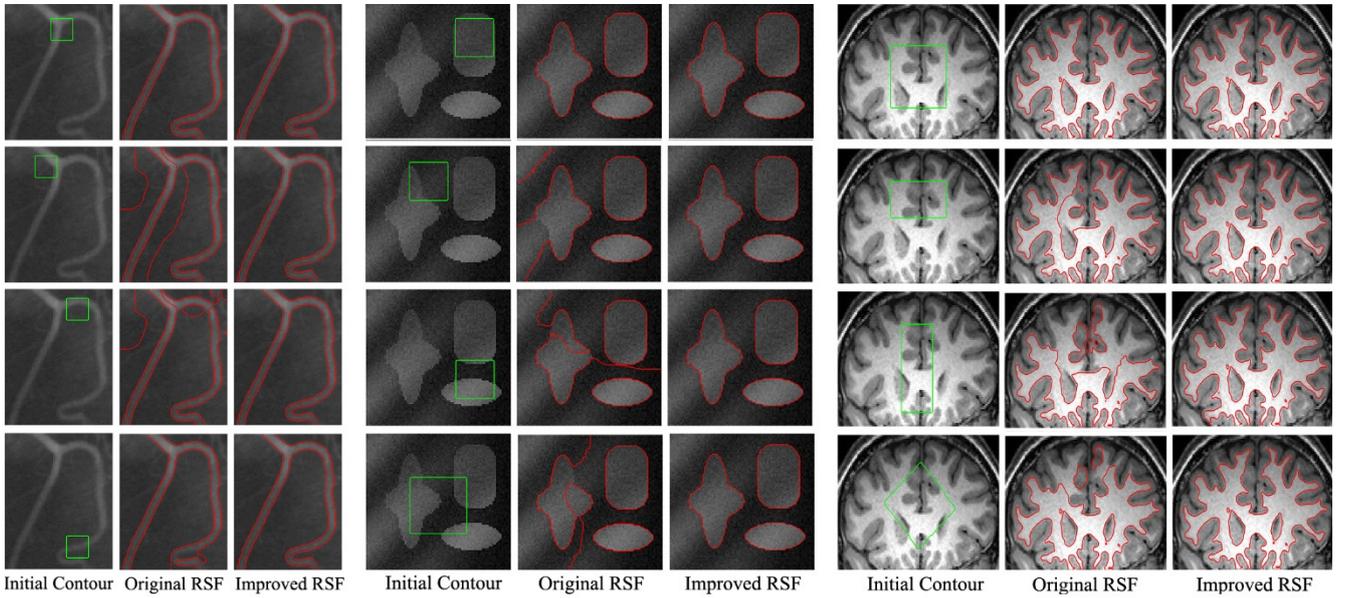

**Fig. 4.** *The segmentation results of original RSF model and improved RSF model with five different initial contours for three images.*

The results in Fig. 4 have demonstrated that the change of initial contour has no effect on the segmentation results in the improved RSF model. But in the original RSF model, it may cause a wrong segmentation result. Thus, some parameters, such as the weight of length term or the scale parameter, need to be adjusted in order to obtain a desired result. Obviously, this is a cumbersome process, and largely depends on human experience. Moreover, the desired result cannot be obtained just by adjusting the parameters for some images. But in the improved RSF model, the satisfactory segmentation result can be obtained without changing any conditions including the parameters. Therefore, the improved RSF model is more robust to initial contour.

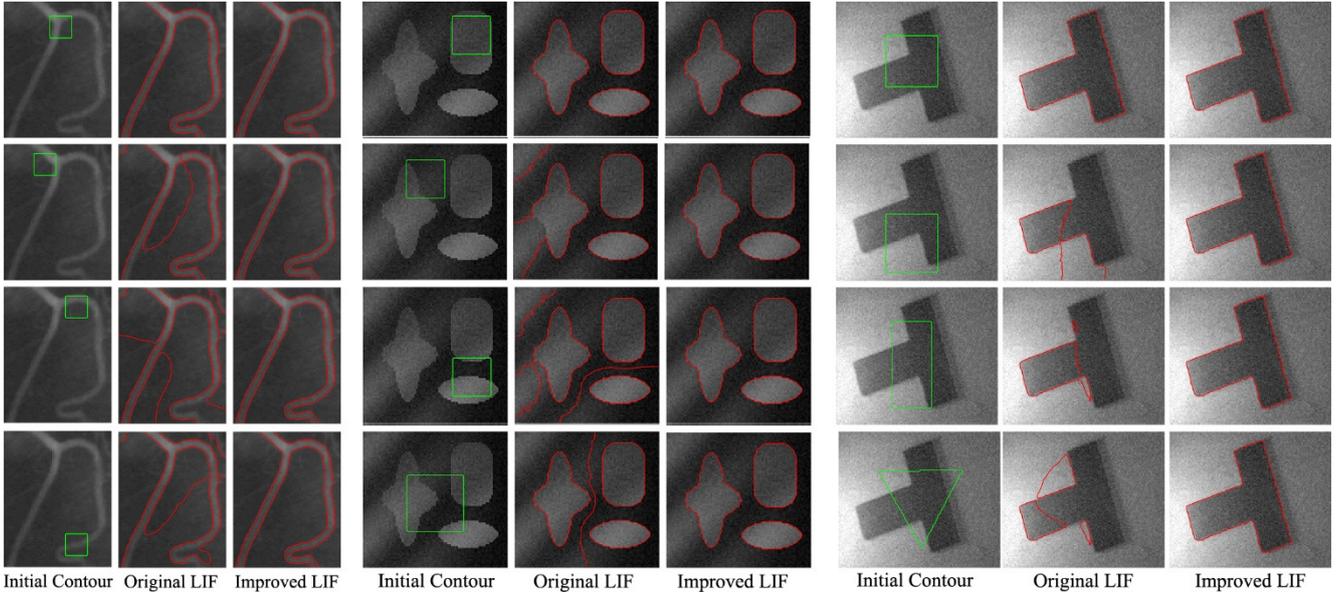

**Fig. 5.** *The segmentation results of original LIF model and improved LIF model with five different initial contours for three images.*

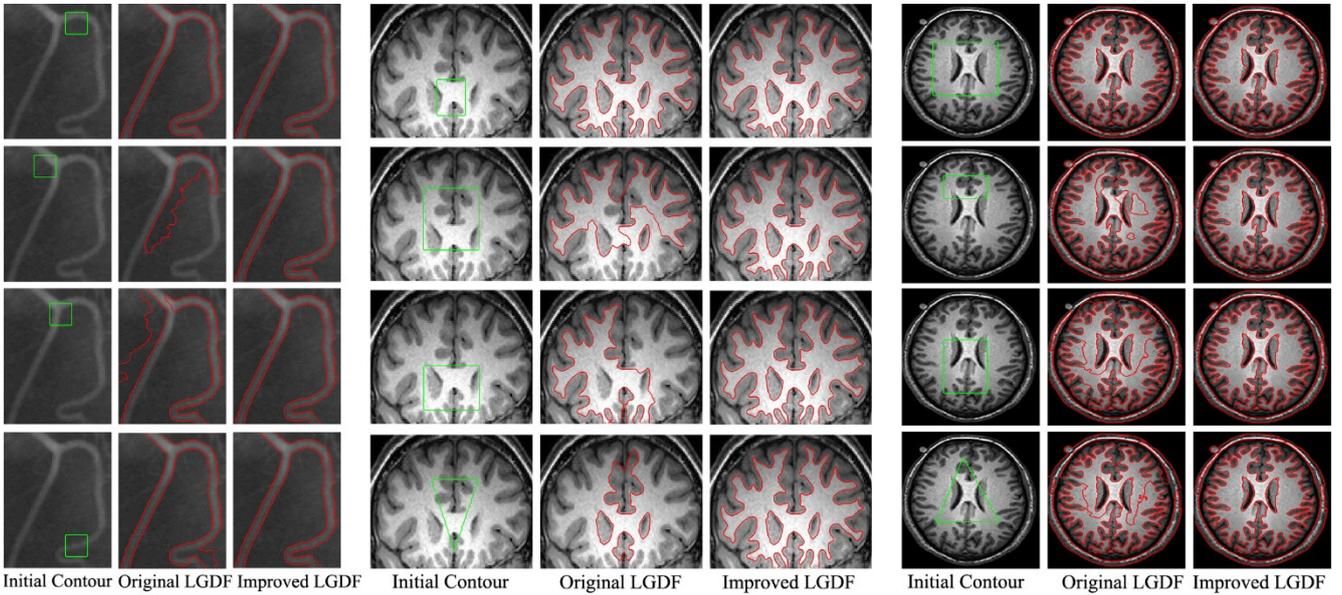

**Fig. 6.** *The segmentation results of original LGDF model and improved LGDF model with five different initial contours for three images.*

Fig. 5 shows the segmentation results of original LIF model and improved LIF model with different initial contours. Fig. 6 shows the segmentation results of original LGDF model and improved LGDF model with different initial contours. Their results are similar to the results in Fig. 4. In the improved LIF and LGDF model, the desired results can be obtained directly under each initial contour. Note, the length term $v = 0.6$ for the right image in Fig. 6.

In summary, the improved models are more robust to the initial contour than the original models because the proposed method can ensure the curve evolution with same direction inherently at each iteration. Fig. 4, Fig. 5 and Fig. 6 have demonstrated that the proposed method can enhance the robustness of initialization.

### 4.3 Multi-phase segmentation results

Then, we take the multi-phase RSF (MRSF) model as an example to test the proposed method for multiple level set segmentation. A brain magnetic resonance (MR) images from McGill Brain Web [26] [25] is chosed as test objects, we need to extract the white matter and gray matter from the background. In practice, because MRSF is much sensitive to the the initial contour, a preliminary segmentation is often used to initialize the contour, such as the simple threshold method. In the experiment, we set three different initial contours, where, the first initial contour is obtained by twice thresholding, as shown in the first row of Fig. 7.

Fig. 7 shows the segmentation results of original MRSF model and improved MRSF model with different initial contours for a brain MR image. Note, the length term $v = 0.003 \times 255^2$ for this image. Under the first initial contour acquired by thresholding, both the original and improved MRSF model obtain a desired result and extract the white matter and gray matter accurately. But under the second and third initial contours, the results of original MRSF model are unsatisfactory, and the extraction of white matter and gray matter are obvious imprecise. In the improved MRSF model, the results are desirable, which are much similar to the result under first initial contour.

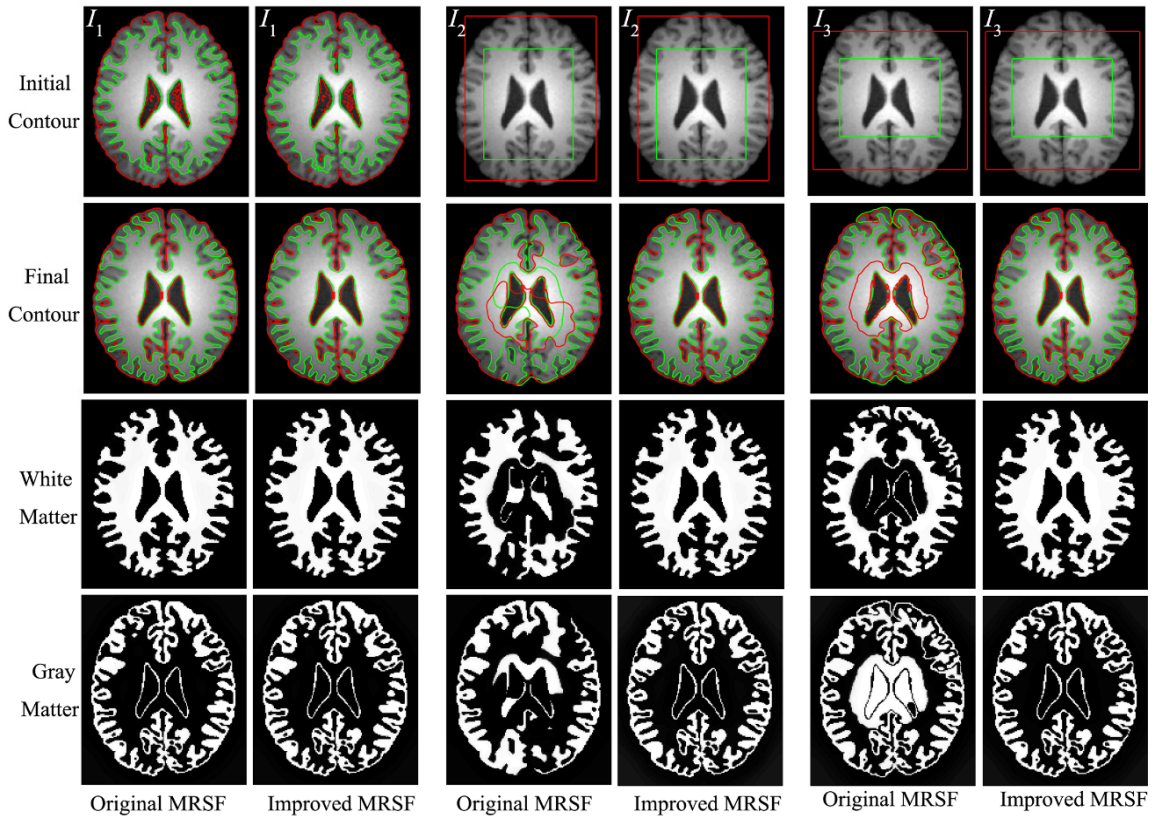

**Fig. 7.** *The segmentation results of original MRSF model and improved MRSF model with different initial contours for a brain MR image.*

Table 1 shows the dice similarity coefficient (DSC) [26] of the segmentation results in Fig. 7. The closer the DSC values to 1, the higher the accuracy of segmentation. From Table 1, we can conclude that the improved MRSF model has extracted the white matter and gray matter precisely while the original MRSF model cannot work well when using the second and third initial contours. Therefore, the improved MRSF model is more robust than original MRSF model in terms of segmenting multi-phase images.

**Table 1** The dice similarity coefficient (DSC) of the white matter / gray matter.

| Initial Contour | Original MRSF | Improved MRSF |
| --- | --- | --- |
| $I_1$ | 0.980 / 0.965 | 0.980 / 0.965 |
| $I_2$ | 0.524 / 0.486 | 0.966 / 0.942 |
| $I_3$ | 0.485 / 0.398 | 0.962 / 0.939 |

## 5 Discussions

### 5.1 About segmentation efficiency

The proposed method is efficient, which hardly increase the segmentation time compared to original models. Table 2 shows the segmentation time under the first initial contours for the images in Fig. 4, Fig. 5 and Fig. 6. Note, the number of iterations is the same in the original and improved models. All these models are implemented in Matlab R2013a on a 2.6-GHz Inter(R) Core(TM) i5 personal computer. According to the experimental data in Table 2, we can know that the time cost of improved models is close to that of the original models.

### 5.2 About the parameter σ

The scale parameter $\sigma$ plays an important role in the local fitting-based models, because it controls the size of local region. In general, the scale parameter $\sigma = 3.0$, which has been used for all the images in this paper. A reasonable large value of $\sigma$ can reduce the dependence of initial contour, but decrease the segmentation accuracy especially when the phenomenon of intensity inhomogeneity is severe [8]. Take the RSF model for example, we set $\sigma = 3, 4, 5$ and 10, respectively, and keep other parameters unchanged. Fig. 8 shows the segmentation results with the change of the scale parameter $\sigma$ for a vessel image. In the original RSF model, the results are wrong when $\sigma = 3$ and 4. Until $\sigma = 5$, the results are correct, but not accurate enough. With the increase of $\sigma$, the results become more inaccurate (i.e., the result when $\sigma = 10$). In the improved RSF model, the correct result can be obtained when $\sigma = 3, 4$ and 5. It means using the proposed method can enhance the robustness of scale parameter $\sigma$. Thus, to obtain a more accurate segmentation result, we can set the scale parameter $\sigma$ to be a relatively small value in the improved models.

**Table 2** Comparisons on segmentation time between the original models and improved models.

| Image | Fig. 4 Original / Improved RSF | Fig. 5 Original / Improved LIF | Fig. 6 Original / Improved LGDF |
|---|---|---|---|
| Left | 1.722 / 1.730 | 1.653 / 1.663 | 2.928 / 2.960 |
| Middle | 0.962 / 0.968 | 0.850 / 0.844 | 4.072 / 4.106 |
| Right | 3.251 / 3.274 | 1.031 / 1.045 | 6.451 / 6.484 |

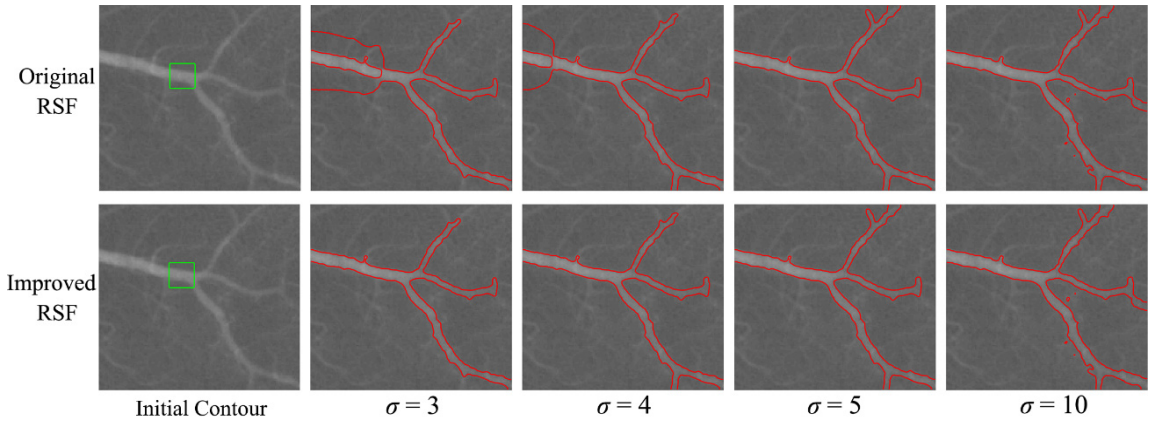

**Fig. 8.** *The segmentation results of original and improved RSF model with the change of the scale parameter σ for a vessel image.*

## 6  Conclusions

In this paper, we have presented a simple method to improve the robustness of initialization for the active contour models driven by local region fitting energy. For the fitting functions on the two sides of curve $f_1$ and $f_2$, we used $min(f_1, f_2)$ and $max(f_1, f_2)$ (or $max(f_1, f_2)$ and $min(f_1, f_2)$) to replace original $f_1$ and $f_2$ to make the whole curve evolves along the same direction. We applied the proposed method to orginal local fitting-based models, including RSF, LIF and LGDF models. Experiments have proved that using the proposed method can enhance the robustness of initial contour and keep the original advantages at the same time.

## 7  Acknowledgements

This work is supported by the National Nature Science Foundation of China [grant numbers 61473201].